\documentclass[runningheads]{llncs}

\usepackage{eccv}

\usepackage{eccvabbrv}

\usepackage{graphicx}
\usepackage{amsmath}
\usepackage{amssymb}

\usepackage{comment}

\usepackage{multirow}

\usepackage{array}
\usepackage{booktabs, makecell}

\usepackage{algorithm}
\usepackage{algpseudocode}

\usepackage[accsupp]{axessibility}  

\usepackage{hyperref}

\usepackage{orcidlink}

\newcolumntype{L}[1]{>{\raggedright\arraybackslash}m{#1}}
\newcolumntype{C}[1]{>{\centering\arraybackslash}m{#1}}
\newcolumntype{R}[1]{>{\raggedleft\arraybackslash}m{#1}}
\newcolumntype{+}{>{\global\let\currentrowstyle\relax}}
\newcolumntype{^}{>{\currentrowstyle}}

\newcommand{\RomNum}[1]{\MakeUppercase{\romannumeral #1}}

\newcommand{\bfr}{\ensuremath{{\mathbf{r}}}}

\newcommand{\bfx}{\ensuremath{{\mathbf{x}}}}

\newcommand{\bfu}{\ensuremath{{\mathbf{u}}}}

\newcommand{\bfU}{\ensuremath{{\mathbf{U}}}}

\usepackage[capitalize]{cleveref}
\crefname{section}{Sec.}{Secs.}
\Crefname{section}{Section}{Sections}
\Crefname{table}{Table}{Tables}
\crefname{table}{Tab.}{Tabs.}

\begin{document}

\title{OMR: Occlusion-Aware Memory-Based Refinement for Video Lane Detection}


\titlerunning{OMR}

\author{Dongkwon Jin\inst{1,2}\orcidlink{0000-0002-6748-3284} \and
Chang-Su Kim\inst{1}\orcidlink{0000-0002-4276-1831}}

\authorrunning{D. Jin and C.-S. Kim}

\institute{School of Electrical Engineering, Korea University, Seoul, Korea \and
Samsung Advanced Institute of Technology \\
\email{dongkwonjin@mcl.korea.ac.kr, changsukim@korea.ac.kr}}

\maketitle

\begin{abstract}
  A novel algorithm for video lane detection is proposed in this paper. First, we extract a feature map for a current frame and detect a latent mask for obstacles occluding lanes. Then, we enhance the feature map by developing an occlusion-aware memory-based refinement (OMR) module. It takes the obstacle mask and feature map from the current frame, previous output, and memory information as input, and processes them recursively in a video. Moreover, we apply a novel data augmentation scheme for training the OMR module effectively. Experimental results show that the proposed algorithm outperforms existing techniques on video lane datasets. Our codes are available at \href{https://github.com/dongkwonjin/OMR}{https://github.com/dongkwonjin/OMR}.
  \keywords{Video lane detection \and Occlusion \and Feature refinement}
\end{abstract}

\begin{figure}[t]
  \centering
  \includegraphics[width=1\linewidth]{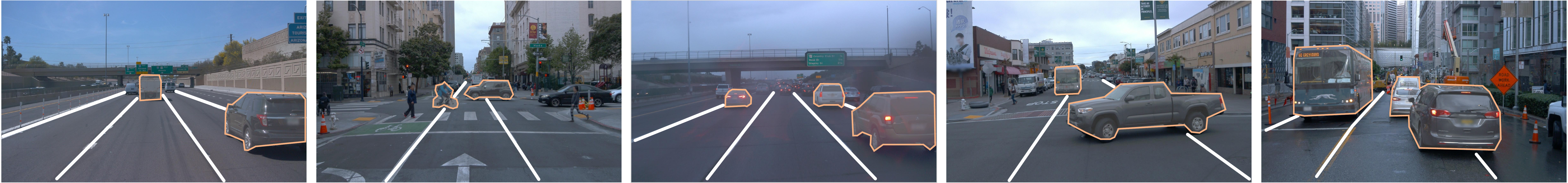}
  \caption{Examples of road scenes, in which some lane parts are occluded by several objects. Visible lanes and obstructing objects are depicted by white lines and orange polygons, respectively.}
  \label{fig:challenging}

\end{figure}

\section{Introduction}
Lane detection aims to localize lanes in a road scene, which is essential for either enabling autonomous driving or assisting human driving. It is, however, difficult to detect lanes, which may be unobvious due to occlusions by nearby vehicles or severe weather conditions. For lane detection, early methods tried to find visible lane cues by extracting low-level features \cite{he2004,aly2008,hillel2014,zhou2010}. Recently, many techniques have been developed to deal with implied lanes using deep features. Some adopt the semantic segmentation framework  \cite{pan2018,zheng2021,qiu2022mfialane,hou2019road,hou2020inter} and classify each pixel into either the lane category or not. Several attempts have been made to extract continuous lane information, including curve modeling \cite{neven2018,wang2020poly,tabelini2021ICPR,liu2021end,feng2022} and keypoint association \cite{qu2021,wang2022,xu2022}. Meanwhile, anchor-based lane detectors \cite{li2019line,tabelini2021CVPR,jin2022,zheng2022,xiao2023} have been proposed. They predefine a set of lane anchors and then detect lanes through the classification and regression of each anchor, ensuring lane continuity. However, all these methods are image-based detectors that process each frame independently, so they often fail to yield temporally stable detection results, especially when some lanes are occluded by objects, as illustrated in Fig.~\ref{fig:challenging}.

\begin{figure}[t]
  \centering
  \includegraphics[width=1\linewidth]{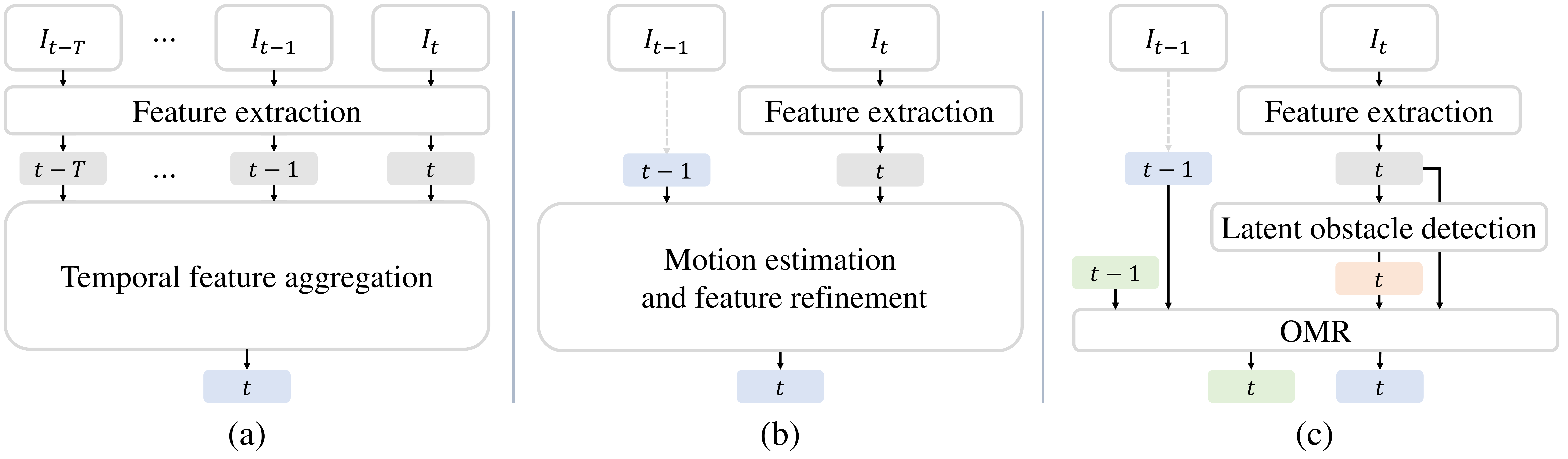}
  \caption{There are three approaches to video lane detection. In (a), the feature maps of a current frame $I_t$ and the past $T$ frames are extracted and mixed to refine the feature map of $I_t$. In (b), only a single previous frame is used to enhance the feature map of $I_t$, and the enhanced one is passed recursively to the subsequent frame. The proposed algorithm in (c) utilizes obstacle and memory information to improve the feature map of $I_t$ via the OMR module. Note that gray, blue, green, and orange boxes represent intra-frame features, refined features, recorded memory, and a latent obstacle mask, respectively.}
  \label{fig:approach}

\end{figure}

Video lane detectors also have been developed. These techniques exploit past information to detect lanes in a current frame, which may help to identify implied lanes more reliably. Most of them \cite{zou2019robust,zhang2021lane,zhang2021,tabelini2022,wang2022video} adopt the framework in Fig.~\ref{fig:approach}(a). These video detectors extract the features of several past and current frames, aggregate those features, and detect lanes in the current frame using the mixed features. However, they do not reuse the mixed features in future frames. Recently, a recursive video lane detector (RVLD) \cite{jin2023} was proposed. As in Fig.~\ref{fig:approach}(b), RVLD enhances the features of a current frame using the single previous frame only through motion estimation and feature refinement. Also, it passes the state of the current frame recursively to the next frame. RVLD outperforms existing image and video lane detectors, but it may detect lanes inaccurately because it heavily relies on the information in a current frame. In particular, when lanes in a current frame are severely occluded by nearby vehicles, RVLD tends to produce unreliable detection results.

In this paper, we propose a novel video lane detector incorporating an occlusion-aware memory-based refinement (OMR) module. As in Fig.~\ref{fig:approach}(c), it utilizes a latent obstacle mask and memory information to enhance the feature map of a current frame. First, we extract a feature map and detect latent obstacles, hindering lane visibility, from the current frame. Then, we refine the feature map through the OMR module, which takes the obstacle mask and feature map from the current frame, the previous output, and the memory information as input. Moreover, we develop an effective data augmentation scheme for training the OMR module robustly. Experimental results demonstrate that the proposed algorithm outperforms existing techniques on both VIL-100 \cite{zhang2021} and OpenLane-V \cite{jin2023} datasets.

This work has the following major contributions:
\begin{itemize}
\itemsep0mm
\item The proposed OMR module improves lane detection results in a current frame by exploiting an obstacle mask and memory information.
\item We introduce a novel training strategy for video lane detection to identify lanes more robustly.
\item The proposed algorithm yields outstanding lane detection results on video datasets.
\end{itemize}

\section{Related Work}

\subsection{Image-Based Lane Detection}
Various techniques have been developed to detect lanes in a still image. Some are based on semantic segmentation \cite{pan2018,zheng2021,qiu2022mfialane,hou2019road,hou2020inter}, in which each pixel is dichotomized into the lane category or not. To boost the pixelwise classification, Pan \etal \cite{pan2018} propagated the information of pixels spatially. In \cite{zheng2021,qiu2022mfialane}, recurrent or multi-scale feature aggregation was performed. Hou \etal \cite{hou2019road} performed self-attention distillation, while Hou \etal \cite{hou2020inter} employed teacher and student networks. For efficient lane detection, Qin \etal \cite{qin2020} determined the location of each lane on selected rows only. Liu \etal \cite{liu2021condlanenet} developed a conditional lane detection scheme based on the row-wise approach.

Several methods attempt to maintain lane continuity by regressing curve parameters \cite{neven2018,wang2020poly,tabelini2021ICPR,liu2021end,feng2022} or associating multiple keypoints \cite{qu2021,wang2022,xu2022}. Neven \etal \cite{neven2018} did the polynomial fitting of segmented lane pixels. In \cite{wang2020poly,tabelini2021ICPR}, polynomial coefficients of lanes were regressed using neural networks. Also, Liu \etal \cite{liu2021end} predicted cubic lane curves based on a transformer network. Feng \etal \cite{feng2022} employed Bezier curves. In \cite{qu2021}, Qu \etal extracted multiple keypoints and linked them to reconstruct lanes. Wang \etal \cite{wang2022} estimated the offsets from a starting point to keypoints and grouped them into a lane instance. Xu \etal \cite{xu2022} predicted four offsets bilaterally from each lane point to the two nearest ones and the two farthest ones.

Meanwhile, the anchor-based detection framework has been adopted in \cite{chen2019,li2019line,xu2020curvelane,tabelini2021CVPR,jin2022,zheng2022,xiao2023}.
These techniques form lane anchors and then classify and regress each anchor by estimating the lane probability and the positional offset. In \cite{chen2019,xu2020curvelane}, vertical line anchors were employed. In \cite{li2019line,tabelini2021CVPR}, straight line anchors were used to extract global features of lanes.
Zheng \etal \cite{zheng2022} extracted multi-scale feature maps and refined them by aggregating global features of learnable line anchors.
Jin \etal \cite{jin2022} introduced data-driven descriptors called eigenlanes. They generated curved anchors as well as straight ones by clustering all training lanes in the eigenlane space. Xiao \etal \cite{xiao2023} produced a heat map to estimate the starting points and directions of anchors.

\subsection{Video-Based Lane Detection}
There are several video-based lane detectors. Most of them combine the features of a current frame with those of several past frames to detect lanes in the current frame, as in Fig.~\ref{fig:approach}(a). To exploit temporal correlation, Zou \etal \cite{zou2019robust} and Zhang \etal \cite{zhang2021lane} employed recurrent neural networks. Zhang \etal \cite{zhang2021} aggregated features of a current frame and multiple past frames based on the attention mechanism \cite{oh2019,vaswani2017}. Tabelini \etal \cite{tabelini2022} fused global features of lanes in video frames after extracting them using the anchor-based detector in \cite{tabelini2021CVPR}. Wang \etal \cite{wang2022video} modified the feature aggregation module in \cite{zheng2021} to exploit spatiotemporal information in neighboring video frames. However, these video detectors do not reuse the aggregated features in future frames. Recently, Jin \etal \cite{jin2023} developed the RVLD method, which uses only a single previous frame but propagates the state of the current frame to the next frame recursively. As in Fig.~\ref{fig:approach}(b), RVLD estimates a motion field, warps the previous output to the current frame, refines the feature map of the current frame, and passes it to the subsequent frame. Despite promising results, RVLD often misses or incorrectly detects unobvious lanes, especially highly occluded lanes. In contrast, the proposed algorithm effectively processes those lanes by detecting latent obstacles and utilizing both memory information and previous output, as shown in Fig.~\ref{fig:approach}(c).

\begin{figure}[t]
  \centering
  \includegraphics[width=1\linewidth]{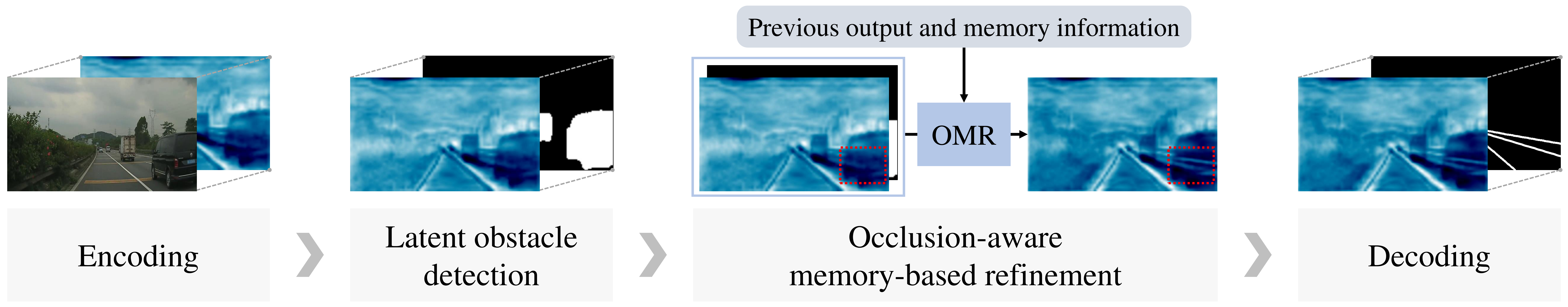}
  \caption{Overview of the proposed algorithm, which performs four steps: encoding, latent obstacle detection, OMR, and decoding. In this example, the rightmost lane is partially occluded by nearby vehicles, so the encoded features are defective, making lane detection difficult. The proposed algorithm, however, can detect the implicit lane precisely by refining the features within the occluded regions effectively. As depicted by dotted red boxes, we see that the proposed OMR module enhances the features of the occluded lane into more discriminative ones.}
  \label{fig:overview}
\end{figure}

\section{Proposed Algorithm}
Given a video sequence, we perform lane detection, which is composed of encoding, latent obstacle detection, feature refinement, and decoding steps. Fig.~\ref{fig:overview} shows an overview of the proposed algorithm. For clarity, we describe the encoding and decoding processes in advance. Notice that the proposed OMR module is performed in the feature refinement step.

\subsection{Encoding}
\label{ssec:encoding}

\begin{figure}[t]
  \centering
  \includegraphics[width=1\linewidth]{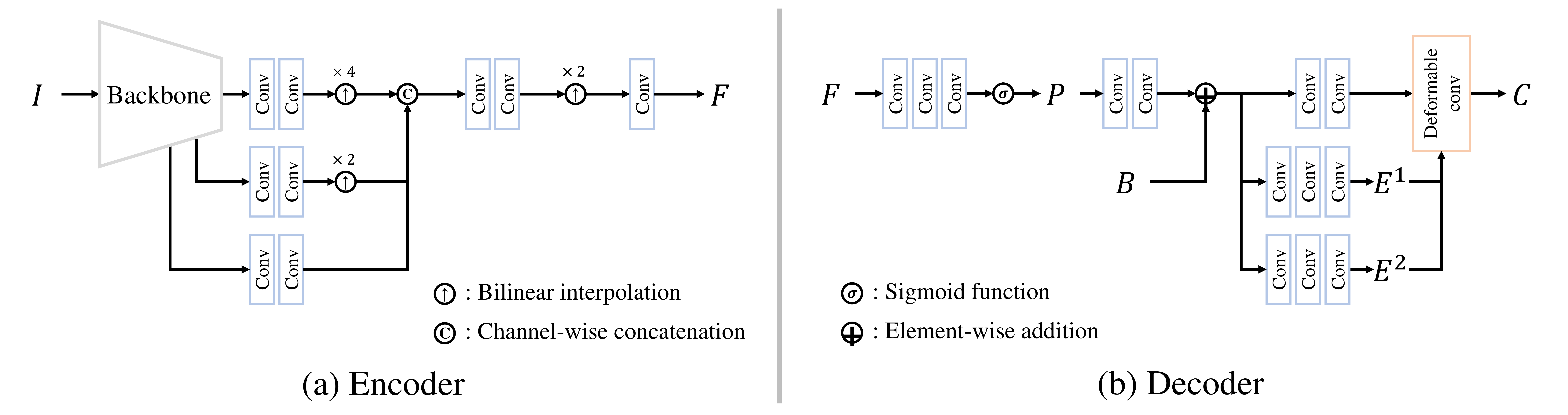}
  \caption{Architecture of the encoder and the decoder: (a) Given an image $I$, three coarsest feature maps are extracted using a backbone network. After matching their channel dimensions and resolutions, they are encoded into a combined feature map $F$. (b) From a feature map $F$, a lane probability map $P$ is estimated. Then, by applying a deformable convolution, a lane coefficient map $C$ is predicted from $P$.}
  \label{fig:enc_dec}
\end{figure}

Given an image $I$, we extract a convolutional feature map $F \in \mathbb{R}^{H\times W\times K}$, as done in \cite{qin2020,jin2022}. Fig.~\ref{fig:enc_dec}(a) shows the encoding process. First, we extract multi-scale feature maps using ResNet18 \cite{he2016deep} as the backbone. Then, we combine the three coarsest maps, which have 1/8, 1/16, and 1/32 of the resolution of $I$, respectively. Specifically, we match the channel dimensions of the feature maps to $K$. We then match the resolutions of the two coarser maps to the finest one via bilinear interpolation and concatenate them. From the concatenated feature map, we obtain $F$ via convolutional and up-sampling layers. We set $K$ to 64.

\subsection{Decoding}
\label{ssec:decoding}
From a feature map $F$, we produce two output maps to determine lanes in $I$, as shown in Fig.~\ref{fig:enc_dec}(b). We obtain a lane probability map $P \in \mathbb{R}^{H\times W\times 1}$ using a series of convolutional layers and a sigmoid function. Let $\bfx$ denote the position vector of a pixel, and let $P(\bfx)$ be the probability that pixel $\bfx$ belongs to a lane. Also, we estimate a lane coefficient map $C \in \mathbb{R}^{H\times W\times M}$. Each element in $C$ is a coefficient vector in the $M$-dimensional eigenlane space \cite{jin2022}, in which lanes are represented compactly with $M$ basis vectors. Since $C(\bfx)$ represents geometric information for a lane containing $\bfx$, we use a positional bias \cite{vaswani2017} to regress the coefficient vector more accurately. To this end, we obtain a sinusoidal positional bias $B \in \mathbb{R}^{H\times W\times K}$ and combine it with $F$ via element-wise addition. From the combined feature map, we generate an offset map $E^1 \in \mathbb{R}^{H\times W\times 50}$, a weight map $E^2 \in \mathbb{R}^{H\times W\times 25}$, and a transformed feature map, and then perform deformable convolution \cite{zhu2019deformable} with a $5 \times 5$ kernel to regress $C$.

Using the probability map $P$ and the coefficient map $C$, we determine reliable lanes through non-maximum suppression (NMS), as done in \cite{jin2023}. First, we select the optimal pixel $\bfx^*$ with the highest probability in $P$. Then, we form the corresponding lane $\bfr$ by linearly combining $M$ eigenlanes with the coefficient vector $C(\bfx^*)$, which is given by
\begin{equation} \label{eq:x_approx}
\bfr = \bfU C(\bfx^*) = [\bfu_1, \cdots, \bfu_M] C(\bfx^*)
\end{equation}
where $\bfu_1, \cdots, \bfu_M$ are the $M$ eigenlanes \cite{jin2022}. Note that $\bfr$ is a column vector containing the horizontal coordinates of lane points, which are uniformly sampled vertically. After dilating the lane curve $\bfr$, we construct a mask and remove the pixels within it to prevent their selection in the remaining iterations. We iterate this NMS process until $P(\bfx^*)$ is higher than 0.5.

Finally, using the selected lanes, we output a lane mask $L \in \mathbb{R}^{H\times W\times 1}$: $L(\bfx) = 1$ if $\bfx$ belongs to a lane, and $L(\bfx)=0$ otherwise.

\begin{figure}[t]
  \centering
  \includegraphics[width=1\linewidth]{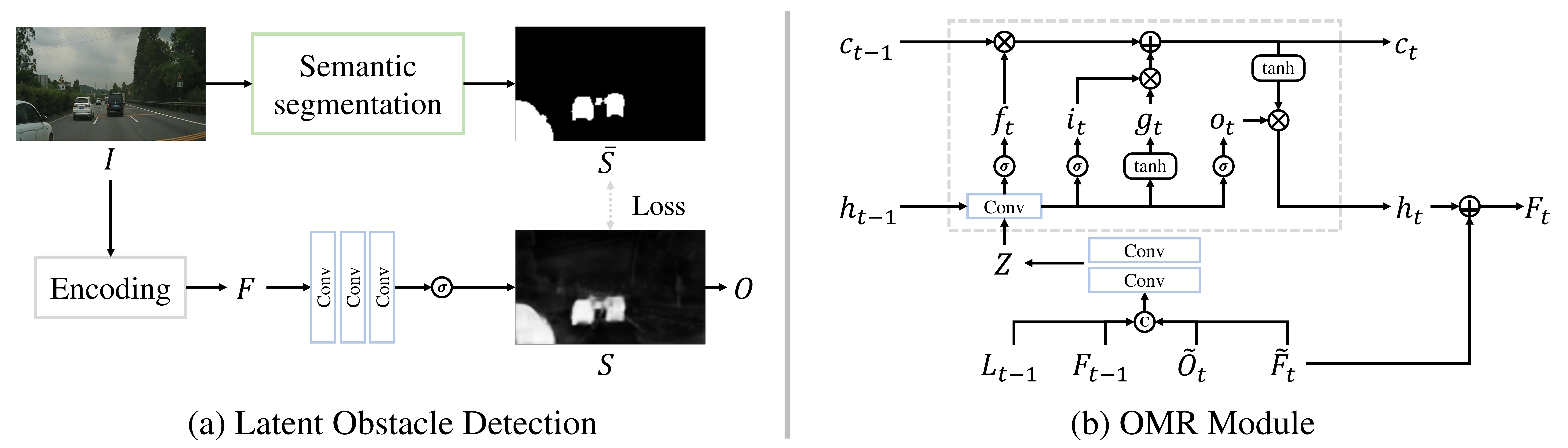}
  \caption{Block diagrams of the latent obstacle detection and OMR: (a) From the encoded feature map $F$, a binary probability map $S$ for latent obstacles is predicted. By thresholding $S$, a binary obstacle mask $O$ is determined. To obtain its ground-truth $\bar{S}$, SegFormer \cite{xie2021segformer}, which is a semantic segmentation algorithm, is employed. (b) In OMR, four input maps $L_{t-1}$, $F_{t-1}$, $\tilde{O}_t$, and $\tilde{F}_t$ are aggregated to $Z$. Then, using the combined feature map $Z$,  ConvLSTM \cite{shi2015} is used to update $(h_{t-1}, c_{t-1})$ to $(h_t, c_t)$ via (\ref{eq:convlstm}). Then, $h_t$ is added to $\tilde{F}_t$ to refine it into $F_t$. Blue boxes represent a series of 2D convolution operations with batch-normalization and ReLU function.}
  \label{fig:LOD_OMRM}
\end{figure}

\subsection{Latent Obstacle Detection}
\label{ssec:inter}

Various objects appear in road environments, such as trucks on highways or pedestrians on crossroads. These objects hinder the visibility of lanes, posing significant challenges in lane detection. To address such occlusion, we treat them as latent obstacles and detect them accordingly. Fig.~\ref{fig:LOD_OMRM}(a) shows the obstacle detection process. More specifically, we predict a binary probability map $S \in \mathbb{R}^{H\times W\times 1}$ from the feature map $F$ of $I$ by
\begin{equation}\label{eq:occ}
    \textstyle
    S = \sigma(w_1(F)),
\end{equation}
where $\sigma$ is the sigmoid function and $w_1$ is composed of 2D convolutional layers. $S(\bfx)$ is the probability that pixel $\bfx$ belongs to an obstacle on a road surface. Then, we obtain an obstacle mask $O \in \mathbb{R}^{H\times W\times 1}$, in which $O(\bfx)$ is assigned to 1 if $S(\bfx)$ is higher than a threshold, and 0 otherwise. We set the threshold to 0.3.

Since there is no ground-truth (GT) segmentation of those obstacles in existing lane datasets, we generate their pseudo-labels by employing a semantic segmentation algorithm. In this work, we adopt SegFormer \cite{xie2021segformer}, which is efficient and yields high performance on the Cityscapes dataset \cite{cordts2016}. Eight of the 19 categories in the dataset, including `car,' `bus,' and `rider' classes, are regarded as potential lane-occluding obstacles. Then, we perform SegFormer to produce a GT binary segmentation mask $\bar{S} \in \mathbb{R}^{H\times W\times 1}$ for those candidates in $I$. Using the predicted map $S$ and its GT mask $\bar{S}$, the obstacle detector in (\ref{eq:occ}) is trained. The training process is detailed in Section \ref{ssec:training}.

\subsection{Occlusion-Aware Memory-Based Refinement}
\label{ssec:inter}

In a current frame $I_t$, some lanes may be unobvious due to the occlusions by neighboring obstacles, as mentioned previously. Furthermore, various factors such as poor lighting and adverse weather conditions affect the visibility of lanes. To deal with these issues, we utilize the obstacle detection results from $I_t$, previous output from $I_{t-1}$, and memory information through the proposed OMR module. Fig.~\ref{fig:LOD_OMRM}(b) shows the structure of the OMR module.

\begin{figure}[t]

    \begin{flushright}
    \setcounter{subfigure}{-1}
    \subfloat[$I_t$] {\includegraphics[width=1.7cm,height=1.1cm]{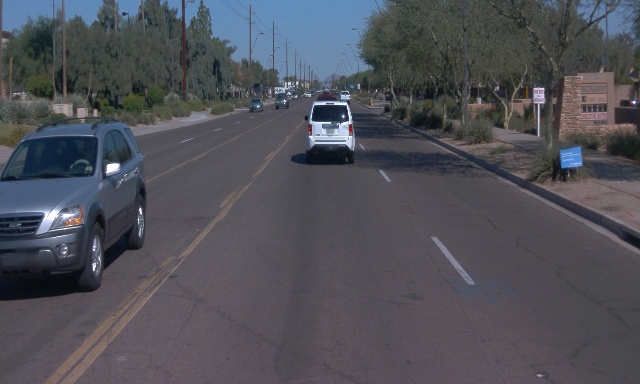}}\,\!\!
    \subfloat[$\tilde{O}_t$] {\includegraphics[width=1.7cm,height=1.1cm]{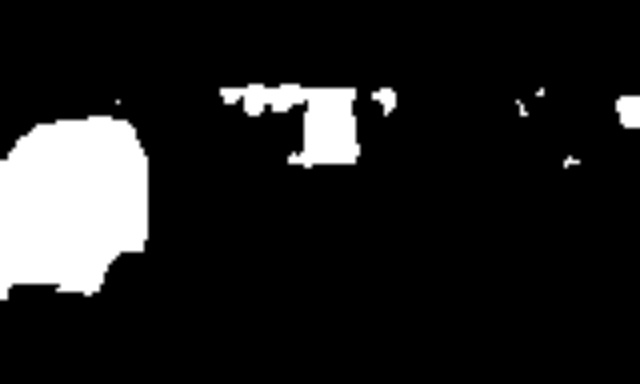}}\,\!\!
    \subfloat[$\tilde{F}_t$] {\includegraphics[width=1.7cm,height=1.1cm]{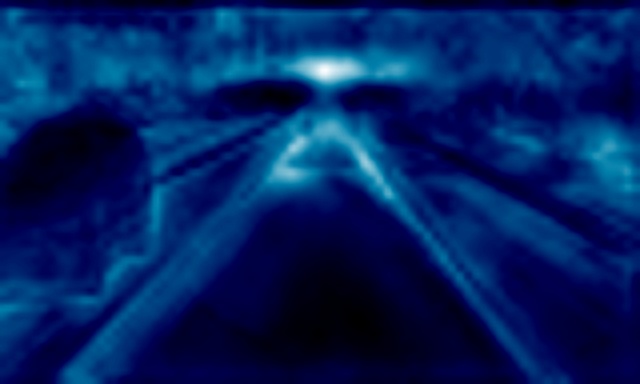}}\,\!\!
    \subfloat[$F_t$] {\includegraphics[width=1.7cm,height=1.1cm]{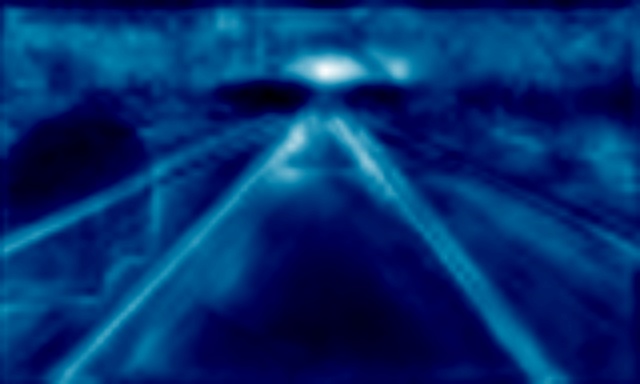}}\,\!\!
    \subfloat[$\tilde{P}_t$] {\includegraphics[width=1.7cm,height=1.1cm]{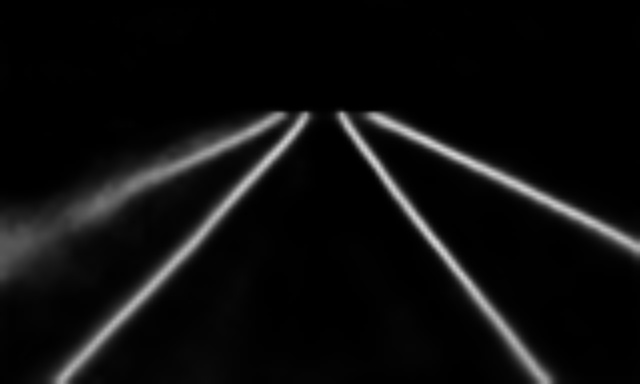}}\,\!\!
    \subfloat[$P_t$] {\includegraphics[width=1.7cm,height=1.1cm]{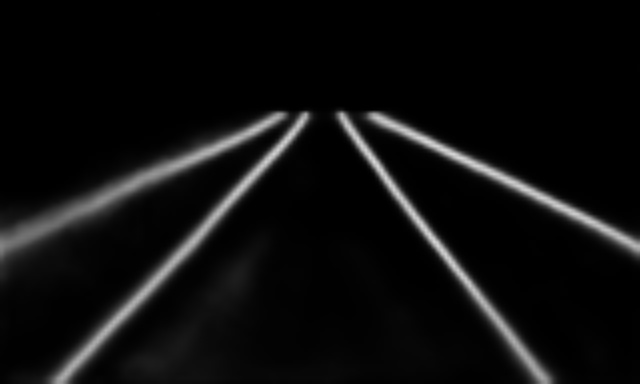}}\,\!\!
    \subfloat[Ground-truth] {\includegraphics[width=1.7cm,height=1.1cm]{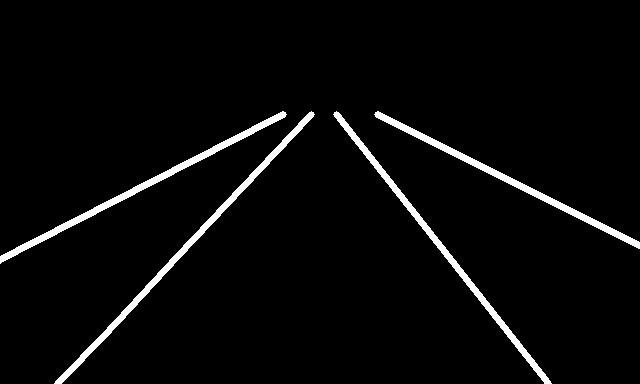}}

    \setcounter{subfigure}{-1}

    \caption
    {
          Visualization of the obstacle mask $\tilde{O}_t$, the feature map $\tilde{F}_t$, the probability map $\tilde{P}_t$, and their enhanced ones $F_t$ and $P_t$. In the current frame $I_t$, some lane parts are occluded by nearby vehicles. The visible lane parts of $\tilde{F}_t$ are sufficiently discriminative for identifying them. In contrast, the features for the occluded parts are not so informative. Thus, $\tilde{P}_t$ is poorly estimated around the occlusions. However, in $F_t$ and $P_t$, the lane features and lane probabilities for the occluded regions are restored faithfully using the proposed OMR module. To visualize these feature maps, min-max normalization is done.
    }
    \label{fig:activation}
    \end{flushright}
\end{figure}

Let $\tilde{F}_t$ be the feature map of $I_t$ obtained by the encoder. In $\tilde{F}_t$ and the following notations, \textit{tilde} represents output produced from a still image. Thus, the probability map $\tilde{P}_t$, the coefficient map $\tilde{C}_t$, the lane mask $\tilde{L}_t$, and the obstacle mask $\tilde{O}_t$ are decoded from $\tilde{F}_t$. Using the OMR module, we aim to refine $\tilde{F}_t$ to $F_t$ and improve the detection results from $\tilde{L}_t$ to $L_t$. To this end, we first obtain a feature map $Z \in \mathbb{R}^{H\times W\times K}$ by aggregating $\tilde{O}_t$ and $\tilde{F}_t$ with the previous output $L_{t-1}$ and $F_{t-1}$ via
\begin{equation}\label{eq:convlstm_in}
    \textstyle
    Z = w_4([w_2(L_{t-1}), F_{t-1}, w_3(\tilde{O}_t), \tilde{F}_t])
\end{equation}
where $[\cdot]$ is channel-wise concatenation, and $w_2$, $w_3$, and $w_4$ are 2D convolution layers. Also, notice that $L_{t-1}$ and $F_{t-1}$ are already refined in the previous step and recursively used in the current step. Then, we exploit memory information by employing a variant of ConvLSTM \cite{shi2015}. Specifically, from the mixed feature map $Z$, we perform a series of ConvLSTM operations by
\begin{equation}\label{eq:convlstm}
\begin{aligned}
    \textstyle
    & f_t = \sigma(w_5(Z) + w_6(h_{t-1})), \\
    & i_t = \sigma(w_7(Z) + w_8(h_{t-1})), \\
    & g_t = \sigma(w_9(Z) + w_{10}(h_{t-1})), \\
    & o_t = \tanh (w_{11}(Z) + w_{12}(h_{t-1})), \\
    & c_t = f_t \odot c_{t-1} + i_t \odot g_t, \;\;\; h_t = o_t \odot c_t.
\end{aligned}
\end{equation}
Here, $w_5, \ldots, w_{12}$ are 2D convolutional layers, and $\odot$ is element-wise multiplication, respectively. Also, $h_t$ and $c_t$ are a hidden state and cell state, and $f_t$, $i_t$, $g_t$, and $o_t$ are a forget gate, input gate, control gate, and output gate, respectively. The four gates are used to update the cell state and hidden state sequentially. $h_1$ and $c_1$ are initialized by learnable parameters. Also, we do not use the cell vectors for estimating the gate parameters. Then, we produce the refined feature map $F_t$ by
\begin{equation}\label{eq:refine}
    \textstyle
    F_t = \tilde{F}_t + h_t,
\end{equation}
Fig.~\ref{fig:activation} illustrates that the OMR module refines $\tilde{F_t}$ to $F_t$ reliably, even though some lane parts are obstructed by vehicles.

\captionsetup[subfigure]{labelformat=empty}
\captionsetup[subfigure]{labelformat=parens}

Lastly, using the refined feature map $F_t$, we produce a reliable lane mask $L_t$ by performing the decoding process, as described in Section \ref{ssec:decoding}.

\subsection{Training}
\label{ssec:training}

\noindent\textbf{Data configuration:}
For each image $I$, we generate a GT lane probability map $\bar{P} \in \mathbb{R}^{H\times W\times 1}$, a GT coefficient map $\bar{C} \in \mathbb{R}^{H\times W\times M}$, and a GT obstacle mask $\bar{S} \in \mathbb{R}^{H\times W\times 1}$. First, $\bar{P}(\bfx) = 1$ if pixel $\bfx$ belongs to a lane, and $\bar{P}(\bfx) = 0$ otherwise. To obtain $\bar{C}$, $M$ eigenlanes are extracted by processing all lanes in a training set, as done in \cite{jin2022}. Then, each lane in an image is transformed to an $M$-dimensional coefficient vector. In the image, $\bar{C}(\bfx)$ is assigned the coefficient vector if $\bfx$ belongs to one of the lanes. Otherwise, if $\bfx$ does not belong to any lane, $\bar{C}(\bfx)$ is assigned the zero vector. To obtain $\bar{S}$, we adopt Segformer \cite{xie2021segformer}, which predicts the semantic segmentation mask for 19 categories. Thus, $\bar{S}(\bfx)$ is set to 1 if $\bfx$ belongs to eight of those categories, such as `car,' `bicycle,' `bus,' `truck,' `train,' `motorcycle,' `person,' and `rider,' and 0 otherwise.

\noindent\textbf{Loss function:}
We perform training in two steps. First, we define the loss for training the encoder and decoder as
\begin{equation}\label{eq:step1}
    \textstyle
    \ell_{\rm step 1} = \ell_{\rm cls}(\tilde{P}, \bar{P})+\ell_{\rm reg}(\tilde{C}, \bar{C})+\ell_{\rm cls}(\tilde{S}, \bar{S}).
\end{equation}
Here, $\tilde{P}$, $\tilde{C}$, and $\tilde{S}$ are the decoded output from the encoded feature map $\tilde{F}$ of $I$. Also, $\ell_{\rm cls}$ is the focal loss \cite{lin2017focal} over binary classes, and $\ell_{\rm reg}$ is the LIoU loss \cite{zheng2022} between a predicted lane contour $\bfr$ in \eqref{eq:x_approx} and its ground-truth $\bar{\bfr}$. Then, we define the loss for training the proposed OMR module as

\begin{equation}\label{eq:step2}
    \textstyle
    \ell_{\rm step 2} = \ell_{\rm cls}(P, \bar{P})+\ell_{\rm reg}(C, \bar{C})
\end{equation}
where $P$ and $C$ are the output from the refined feature map $F$. Also, during the training of the OMR module, we freeze the parameters of the pretrained encoder and decoder.

\begin{figure}[t]
  \centering
  \includegraphics[width=1\linewidth]{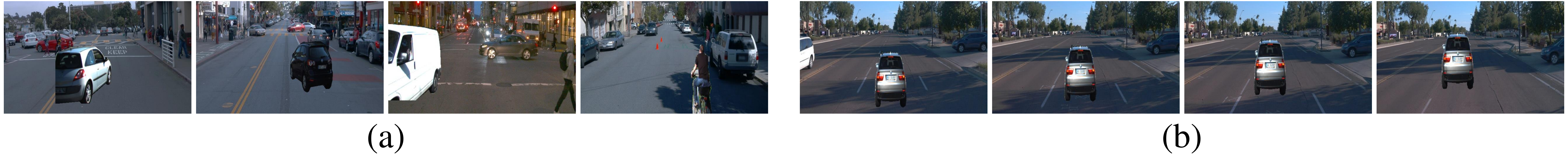}
  \caption{(a) In a training set, each image is synthesized by overlaying new objects, such as vehicles or cyclists, from the KINS dataset. (b) Additionally, video sequences are regenerated by linearly varying the sizes and positions of these objects over frames. The resulting images appear natural because fully shaped objects are extracted from KINS.}
  \label{fig:augmentation}

\end{figure}

\noindent\textbf{Data augmentation:}
In real-world environments, lanes unexpectedly disappear and reappear due to occlusions by nearby objects. To cope with such challenging scenarios reliably, we introduce a data augmentation scheme, which is applied to the training of the OMR module. To this end, we employ the KINS dataset \cite{qi2019amodal}. It is an amodal instance segmentation dataset, in which a fully-shaped mask per each object is given involving its occluding parts. Given an original video sequence, we randomly select an object from the KINS dataset and then attach its full shape to the video frames. We also vary the size and position of the object linearly over frames. Fig.~\ref{fig:augmentation} shows some examples of the synthetically generated frames.

\section{Experimental Results}

\subsection{Implementation Details}

We adopt ResNet18 \cite{he2016deep} as a backbone. We use AdamW optimizer \cite{loshchilov2017} with an initial learning rate of $10^{-1}$ and halve it after every 100,000 iterations four times. Also, we use a batch size of four for 400,000 iterations. We resize an input image to $384 \times 640$. As the default setting, we set $H=96$, $W=160$, $K=64$, and $M=6$. We also employ SegFormer-B5 \cite{xie2021segformer} for the semantic segmentation.

\subsection{Datasets}

VIL-100 \cite{zhang2021} is the first dataset for video lane detection containing 100 videos. It is split into 80 training and 20 test videos. Each video has 100 frames. VIL-100 includes some challenging scenes in which some lanes are highly occluded by big trucks or buses. In each frame, 2D lane coordinates up to 6 road lanes are annotated.

OpenLane-V \cite{jin2023}, which is modified from OpenLane \cite{chen2022},  is a huge and diverse video lane dataset. It consists of about 90K images from 590 videos. It is split into a training set of 70K images from 450 videos and a test set of 20K images from 140 videos. As in the CULane dataset \cite{pan2018}, up to 4 road lanes are annotated in each image, corresponding to ego and alternative lanes. OpenLane-V is more difficult for lane detection than VIL-100, because of various challenging factors: lane occlusions, severe weather conditions, poor illumination, or lack of lane marking on crossroads.


\begin{table*}[t]\centering
    \renewcommand{\arraystretch}{0.88}
    \caption
    {
        Comparison of mIoU, $\rm F1$ scores, flickering, and missing rates on VIL-100: image lane detectors and video ones are listed separately.
    }
    \resizebox{0.88\linewidth}{!}{
    \begin{tabular}[t]{+L{2.5cm}^C{2.0cm}^C{2.0cm}^C{2.0cm}^C{2.0cm}^C{2.0cm}}
    \toprule
    & Approach & mIoU$(\uparrow)$ & ${\rm F1}(\uparrow)$ & ${\rm R}_{\rm F}(\downarrow)$ & ${\rm R}_{\rm M}(\downarrow)$ \\
    \midrule
         LaneNet \cite{neven2018}       & \multirowcell{7}[0pt][c]{Image-based} & 0.633 & 0.721 & - & - \\
         ENet-SAD \cite{hou2019road}    & & 0.616 & 0.755 & - & - \\
         LSTR \cite{liu2021}            & & 0.573 & 0.703 & - & - \\
         RESA \cite{zheng2021}          & & 0.702 & 0.874 & - & - \\
         LaneATT \cite{tabelini2021CVPR}    & & 0.664 & 0.823 & - & - \\
         MFIALane \cite{qiu2022mfialane}& & - & 0.905 & 0.047 & 0.128 \\
         ADNet \cite{xiao2023}          & & \underline{0.781} & 0.920 & 0.039 & \underline{0.043}\\
    \midrule
        MMA-Net \cite{zhang2021}        & \multirowcell{4}[0pt][c]{Video-based} & 0.705 & 0.839 & 0.042 & 0.127 \\
        LaneATT-T \cite{tabelini2022}   & & 0.692 & 0.846 & - & - \\
        TGC-Net \cite{wang2022video}    & & 0.738 & 0.892 & - & - \\
        RVLD \cite{jin2023}             & & \bf{0.787} & \underline{0.924} & \underline{0.038} & 0.050 \\
    \midrule
        Proposed                        & \multirowcell{1}[0pt][c]{Video-based} & 0.774 & \bf{0.936} & \bf{0.026} & \bf{0.038} \\
    \bottomrule
    \end{tabular}}
    \label{table:vil100}
\end{table*}

\subsection{Evaluation Metrics}

\noindent\textbf{Image metrics:}
For lane detection, image-based metrics are generally employed. Each lane is regarded as a thin stripe with a 30-pixel width \cite{pan2018}. Then, a predicted lane is declared correct if its IoU ratio with GT is greater than 0.5. The precision and the recall are computed by
\begin{equation}\label{eq:pre_rec}
    \textstyle
    {\rm Precision} = \frac{\rm TP}{\rm TP + FP}, \;\;\; {\rm Recall} = \frac{\rm TP}{\rm TP + FN}
\end{equation}
where $\rm TP$ is the number of correctly detected lanes, $\rm FP$ is that of false positives, and $\rm FN$ is that of false negatives. Then, the F-measure is defined as
\begin{equation}\label{eq:fscore}
    \textstyle
    {\rm F1} =  \frac{2 \times \rm Precision \times \rm Recall}{\rm Precision + \rm Recall}.
\end{equation}
Also, mIoU is computed by averaging the IoU scores of correctly detected lanes.

\noindent\textbf{Video metrics:}
In autonomous driving systems, achieving temporally stable lane detection is crucial to prevent hazardous situations caused by the sudden detection or absence of a lane within a frame. To assess the temporal stability of detected lanes, two video metrics \cite{jin2023} are employed. There are three cases for a matching pair of lanes at adjacent frames: \textit{Stable}, \textit{Flickering}, and \textit{Missing}. A stable case is one where a lane is detected successfully in both frames. In a flickering case, a lane is detected in one frame but missed in the other. A missing case is the worst one in which both frames miss a lane consecutively.

Let $\rm N$ be the number of GT lanes that have matching instances at previous frames, and let $\rm N_S$, $\rm N_F$, $\rm N_M$ be the numbers of stable, flickering, and missing cases, respectively. Note that $\rm N= N_S +  N_F +  N_M$. Then, the flickering and missing rates are defined as
\begin{equation}\label{eq:pre_rec}
    \textstyle
    \rm
    {\rm R}_{\rm F} = \frac{N_F}{N}, \quad {\rm R}_{\rm M} = \frac{N_M}{N},
\end{equation}
where the IoU threshold for correct detection is 0.5.

\begin{figure}[t]
  \centering
  \includegraphics[width=1\linewidth]{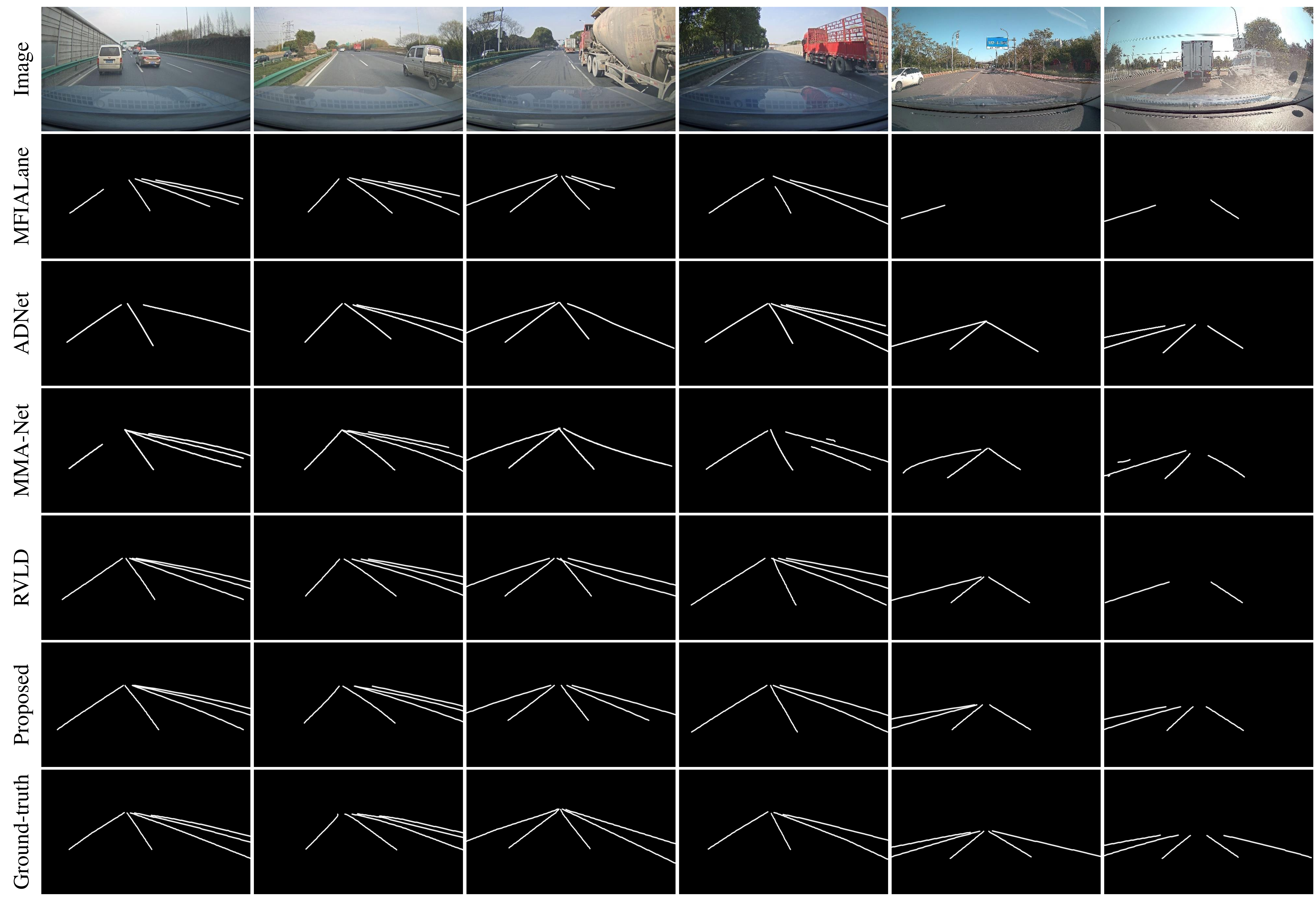}
  \caption{Comparison of lane detection results on the VIL-100 dataset.}

  \label{fig:vil100_result}

\end{figure}

\subsection{Comparative Assessment}

\noindent\textbf{VIL-100:}
We compare the proposed algorithm with conventional image lane detectors \cite{neven2018,hou2019road,liu2021,zheng2021,tabelini2021CVPR,qiu2022mfialane,xiao2023} and video ones \cite{zhang2021,tabelini2022,wang2022video,jin2023} on VIL-100.
Table~\ref{table:vil100} lists the mIoU, $\rm F1$ scores, ${\rm R}_{\rm F}$ rates, and ${\rm R}_{\rm M}$ rates. The proposed algorithm outperforms the existing techniques in every metric, except for mIoU. Especially, the proposed algorithm is better than the state-of-the-art video lane detector RVLD by the same margins of 0.012 in $\rm F1$, ${\rm R}_{\rm F}$, and ${\rm R}_{\rm M}$. RVLD improves the detection results in a current frame using a single previous frame only based on motion estimation and feature refinement. However, it may fail to detect implied lanes occluded by neighboring vehicles. This is because the motion estimator in RVLD tends to produce inaccurate motion field for occluded regions. In contrast, the proposed algorithm detects those lanes more reliably by exploiting the obstacle masks and memory information. ADNet \cite{xiao2023}, a recent image-based detector, yields decent performances, but the scores are inferior to those of the proposed algorithm in most metrics.

Fig.~\ref{fig:vil100_result} presents some detection results. Both MFIALane and ADNet fail to detect unobvious lanes precisely, for it is image-based. MMA-Net does not detect those lanes, even though it uses several past frames as input. RVLD is better than these techniques, but it also processes the occluded lanes poorly. In contrast, the proposed algorithm provides better results based on the obstacle reasoning effectively. 

\begin{figure}[t]
  \centering
  \includegraphics[width=1\linewidth]{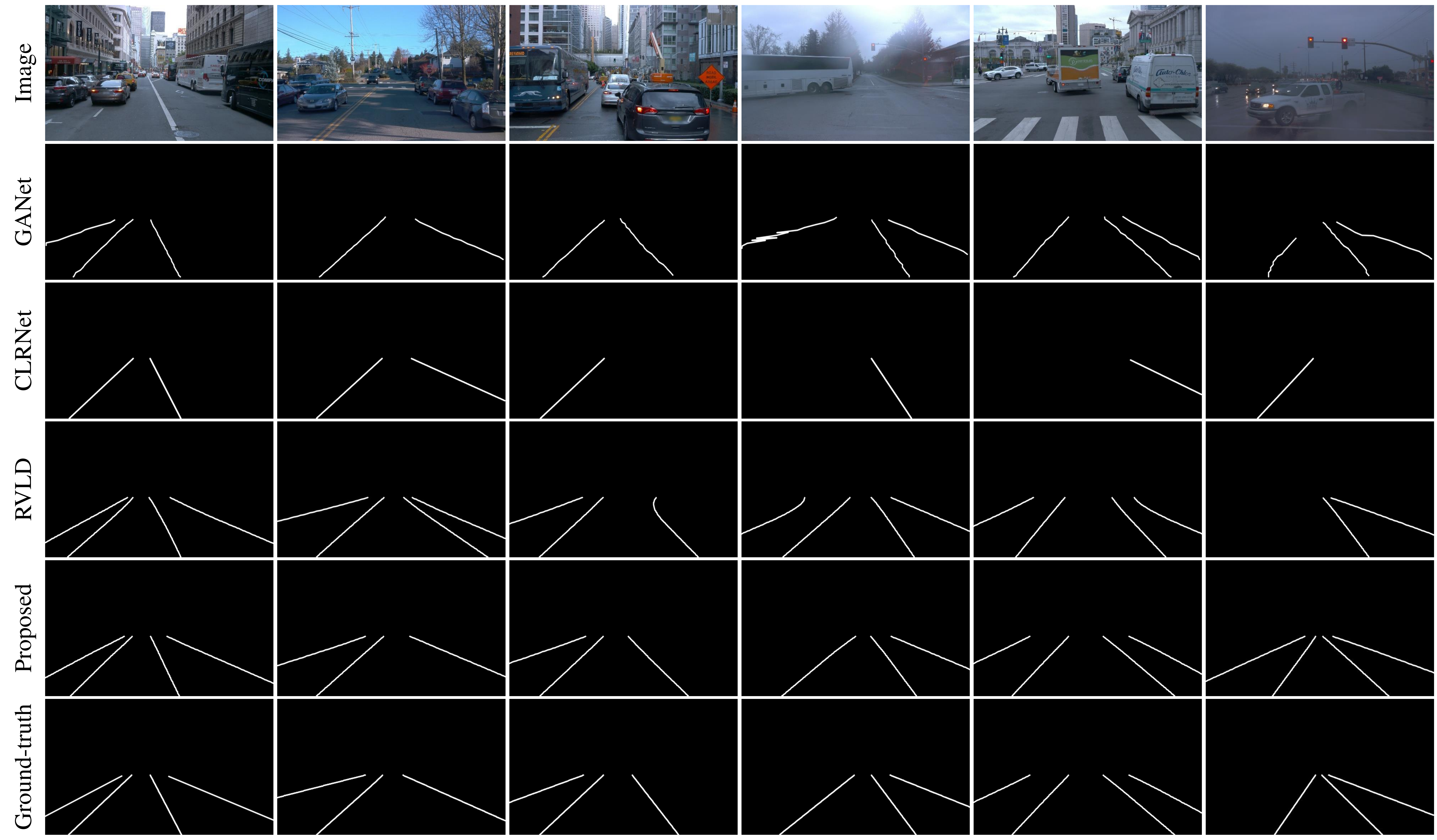}
  \caption{Comparison of lane detection results on the OpenLane-V dataset.}

  \label{fig:openlane_result}
\end{figure}

\begin{table*}[t]\centering
    \renewcommand{\arraystretch}{0.88}
    \caption
    {
        Comparison on OpenLane-V.
    }

    \resizebox{0.88\linewidth}{!}{
    \begin{tabular}[t]{+L{3.0cm}^C{2.0cm}^C{2.0cm}^C{2.0cm}^C{2.0cm}^C{2.0cm}}
    \toprule
    & Approach & mIoU$(\uparrow)$ & ${\rm F1}(\uparrow)$ & ${\rm R}_{\rm F}(\downarrow)$ & ${\rm R}_{\rm M}(\downarrow)$ \\
    \midrule
         MFIALane \cite{qiu2022mfialane}        & \multirowcell{4}[0pt][c]{Image-based} & 0.697 & 0.723 & 0.061 & 0.300 \\
         CondLaneNet \cite{liu2021condlanenet}  & & 0.698 & 0.780 & 0.047 & 0.239 \\
         GANet \cite{wang2022}                  & & 0.716 & 0.801 & 0.048 & 0.198 \\
         CLRNet \cite{zheng2022}                & & \underline{0.735} & 0.789 & 0.054 & 0.224\\
    \midrule
         ConvLSTM \cite{zou2019robust}              &  \multirowcell{4}[0pt][c]{Video-based} & 0.529 & 0.641 & 0.058 & 0.282\\
         ConvGRUs \cite{zhang2021lane}              & & 0.540 & 0.641 & 0.064 & 0.288\\
         MMA-Net \cite{zhang2021}               & & 0.574 & 0.573 & 0.044 & 0.461 \\
         RVLD \cite{jin2023}                   & & 0.727 & \underline{0.825} & \bf{0.014} & \underline{0.167}\\
    \midrule
         Proposed                   & \multirowcell{1}[0pt][c]{Video-based} & \bf{0.742} & \bf{0.836} & \underline{0.016} & \bf{0.162}\\
    \bottomrule
    \end{tabular}}
    \label{table:openlane}

\end{table*}

\noindent\textbf{OpenLane-V:}
Table~\ref{table:openlane} compares the proposed algorithm with the image lane detectors \cite{liu2021condlanenet,qiu2022mfialane,wang2022,zheng2022} and the video lane detectors \cite{zou2019robust,zhang2021lane,zhang2021,jin2023} on OpenLane-V. We see that the proposed algorithm outperforms the existing techniques in most metrics. GANet \cite{wang2022} and CLRNet \cite{zheng2022}, which are recent image-based detectors, perform well in image metrics. But, their flickering rates ${\rm R}_{\rm F}$ and missing rates ${\rm R}_{\rm M}$ are relatively high. RVLD \cite{jin2023} yields the lowest ${\rm R}_{\rm F}$, but it is inferior to the proposed algorithm in other metrics. Specifically, the proposed algorithm outperforms RVLD by margins of 0.015, 0.011, and 0.005 in mIoU, $\rm F1$, and ${\rm R}_{\rm M}$, respectively. Note that reducing ${\rm R}_{\rm M}$ is more critical than reducing ${\rm R}_{\rm F}$ because missing lanes consecutively at both frames represents the worst scenario in video lane detection. These experimental results indicate that the proposed algorithm is temporally more stable than RVLD.

Fig.~\ref{fig:openlane_result} shows detection results. Image-based techniques inaccurately detect implied lanes or simply miss them in challenging scenes. RVLD is better than these detectors but underperforms for highly occluded lanes. In contrast, the proposed algorithm detects those lanes reliably.

\subsection{Ablation Studies}
We conduct ablation studies to analyze the efficacy of the proposed algorithm and its components. Table~\ref{table:ablation} compares several ablated methods on VIL-100. Method \RomNum{1} detects lanes in a current frame $I_t$ without exploiting the latent obstacle mask in the proposed OMR module. In other words, it excludes $\tilde{O}_t$ in (\ref{eq:convlstm_in}). Method \RomNum{2} does not use the memory information by removing the ConvLSTM block in the OMR module. Thus, $Z$ directly becomes $F_t$ in (\ref{eq:refine}). In Method \RomNum{3}, the OMR module is not trained using synthetically generated data. Method \RomNum{4}, the proposed algorithm, applies the OMR module along with the data augmentation scheme.

\begin{table}[t]\centering
    \renewcommand{\arraystretch}{0.88}
    \caption
    {
        Ablation studies of the proposed algorithm on VIL-100.
    }

    \resizebox{0.88\linewidth}{!}{
    \begin{tabular}[t]{+C{0.2cm}^L{1.0cm}^C{2.1cm}^C{2.1cm}^C{2.1cm}^C{1.5cm}^C{1.5cm}^C{1.5cm}}
    \toprule
                    & & Obstacle mask & Memory & Synthetic data & ${\rm F1}$ & ${\rm R}_{\rm F}$ & ${\rm R}_{\rm M}$ \\
    \midrule
        \RomNum{1}  & & 	   & \checkmark &\checkmark        & 0.933 & 0.026 & 0.043 \\
        \RomNum{2}  & & \checkmark	   & & \checkmark          & 0.932 & 0.024 & 0.041 \\
        \RomNum{3}  & & \checkmark & \checkmark &              & 0.929 & 0.036 & 0.043 \\
        \RomNum{4}  & & \checkmark & \checkmark & \checkmark   & 0.936 & 0.026 & 0.038 \\
    \bottomrule
    \end{tabular}}
    \label{table:ablation}

\end{table}

\begin{figure}[t]

    \begin{center}

    \setcounter{subfigure}{-1}

    \subfloat {\includegraphics[width=1.7cm,height=1.1cm]{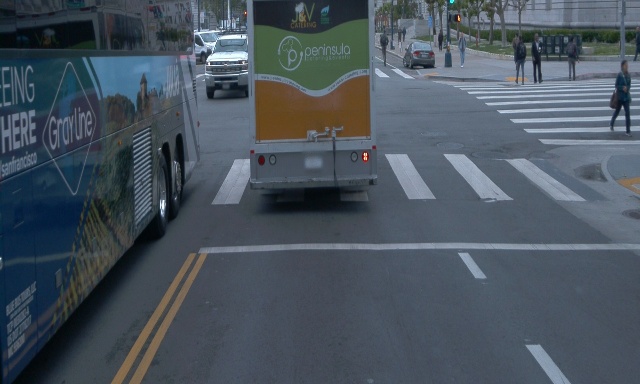}}\,\!\!
    \subfloat {\includegraphics[width=1.7cm,height=1.1cm]{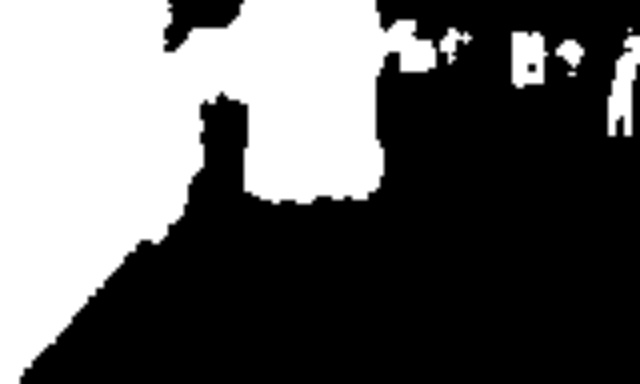}}\,\!\!
    \subfloat {\includegraphics[width=1.7cm,height=1.1cm]{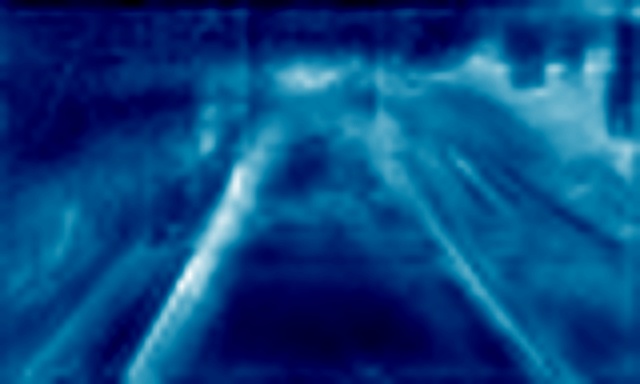}}\,\!\!
    \subfloat {\includegraphics[width=1.7cm,height=1.1cm]{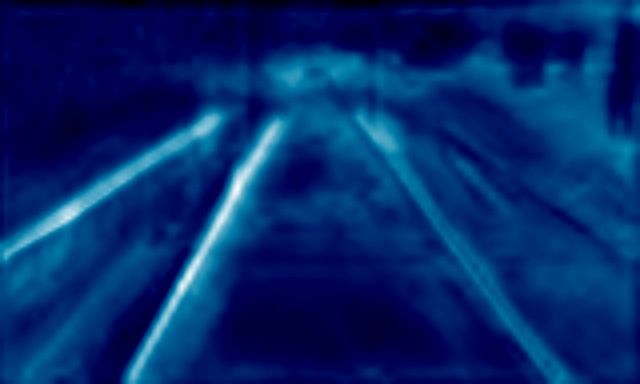}}\,\!\!
    \subfloat {\includegraphics[width=1.7cm,height=1.1cm]{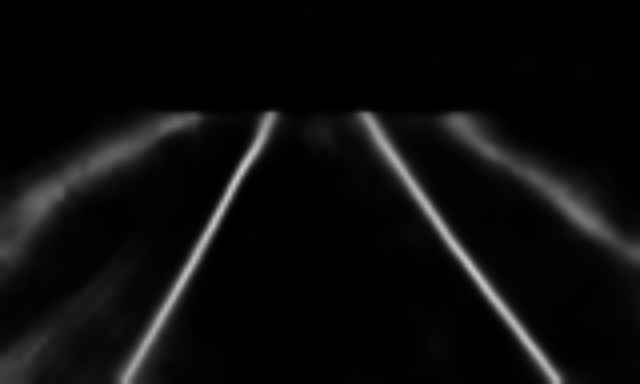}}\,\!\!
    \subfloat {\includegraphics[width=1.7cm,height=1.1cm]{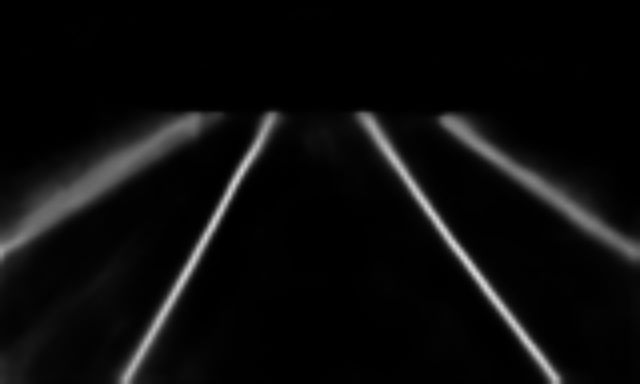}}\,\!\!
    \subfloat {\includegraphics[width=1.7cm,height=1.1cm]{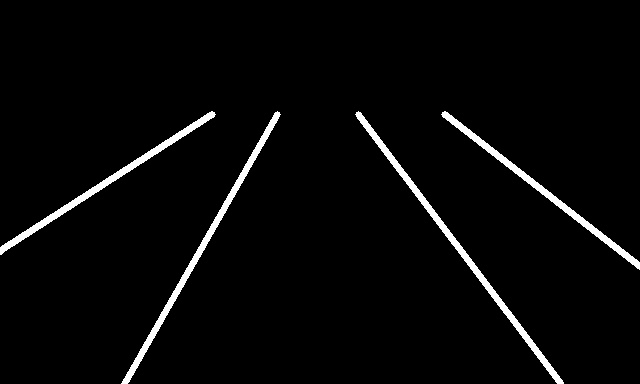}}\\

    \subfloat {\includegraphics[width=1.7cm,height=1.1cm]{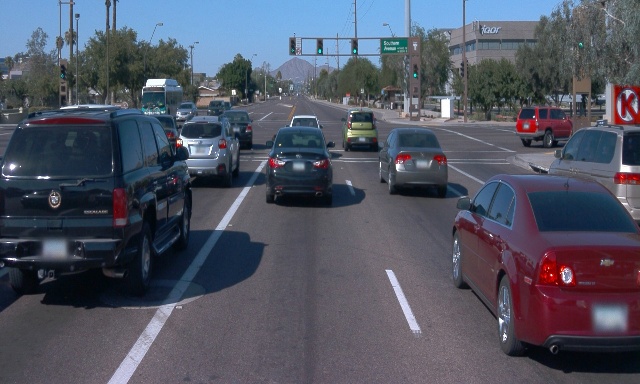}}\,\!\!
    \subfloat {\includegraphics[width=1.7cm,height=1.1cm]{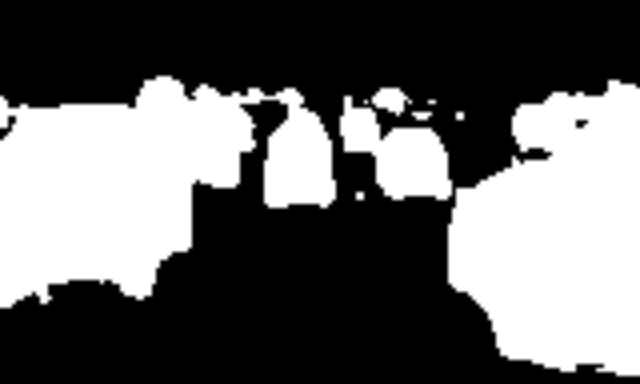}}\,\!\!
    \subfloat {\includegraphics[width=1.7cm,height=1.1cm]{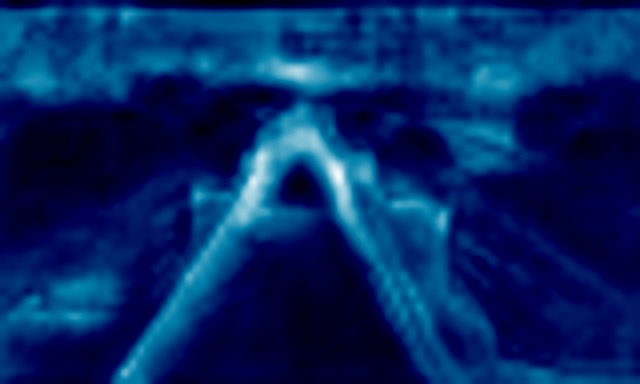}}\,\!\!
    \subfloat {\includegraphics[width=1.7cm,height=1.1cm]{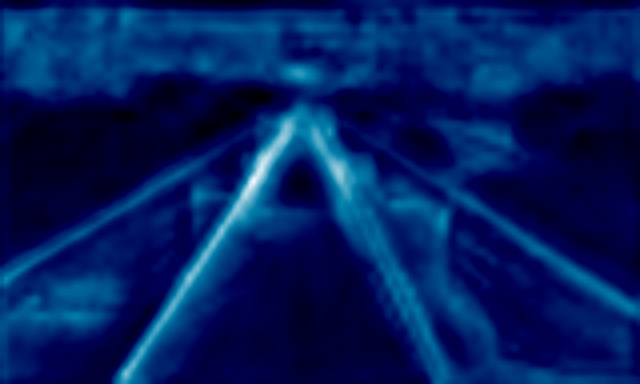}}\,\!\!
    \subfloat {\includegraphics[width=1.7cm,height=1.1cm]{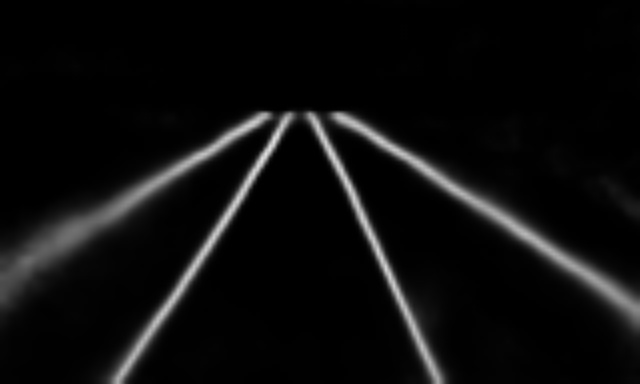}}\,\!\!
    \subfloat {\includegraphics[width=1.7cm,height=1.1cm]{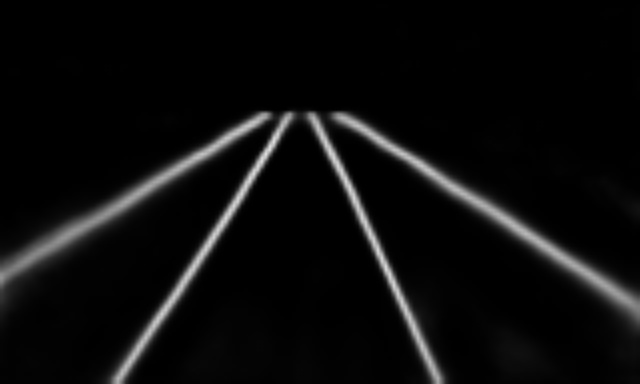}}\,\!\!
    \subfloat {\includegraphics[width=1.7cm,height=1.1cm]{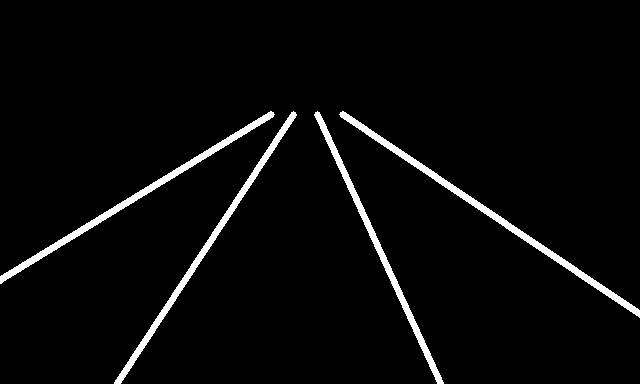}}\\

    \setcounter{subfigure}{-1}

    \subfloat {\includegraphics[width=1.7cm,height=1.1cm]{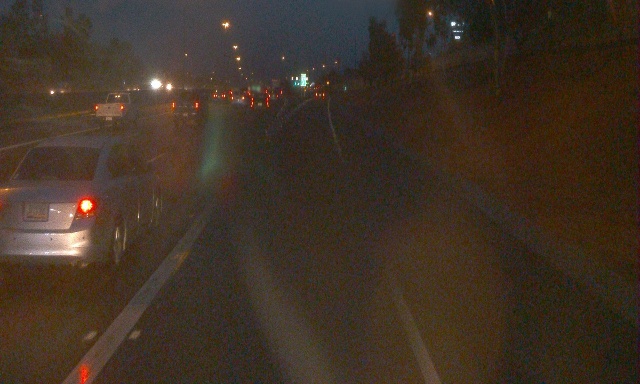}}\,\!\!
    \subfloat {\includegraphics[width=1.7cm,height=1.1cm]{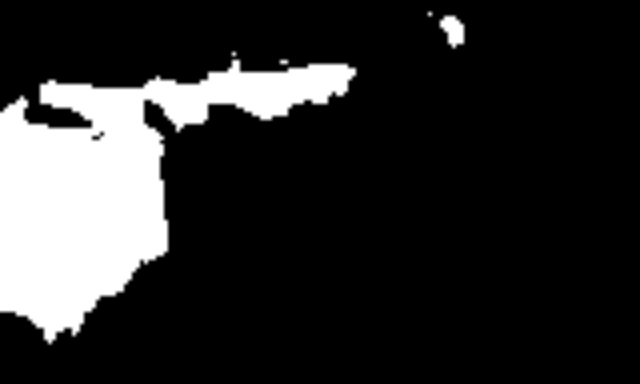}}\,\!\!
    \subfloat {\includegraphics[width=1.7cm,height=1.1cm]{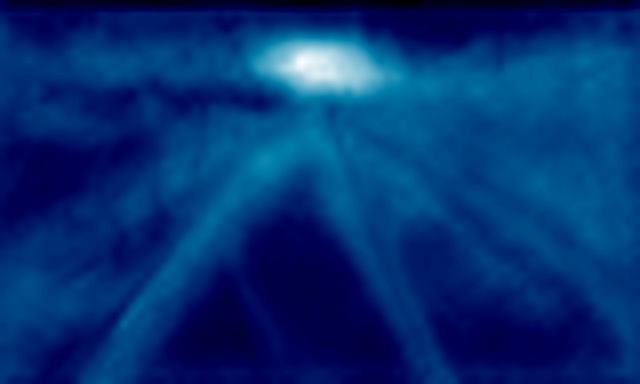}}\,\!\!
    \subfloat {\includegraphics[width=1.7cm,height=1.1cm]{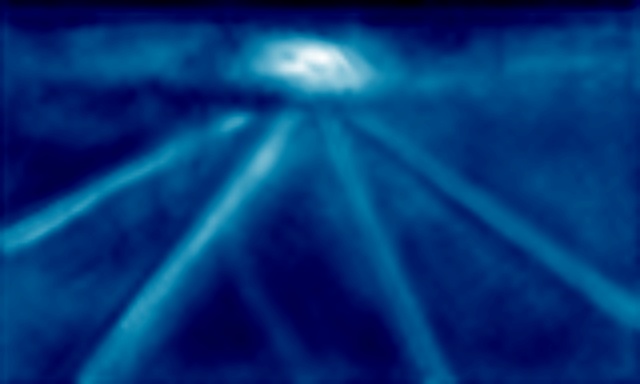}}\,\!\!
    \subfloat {\includegraphics[width=1.7cm,height=1.1cm]{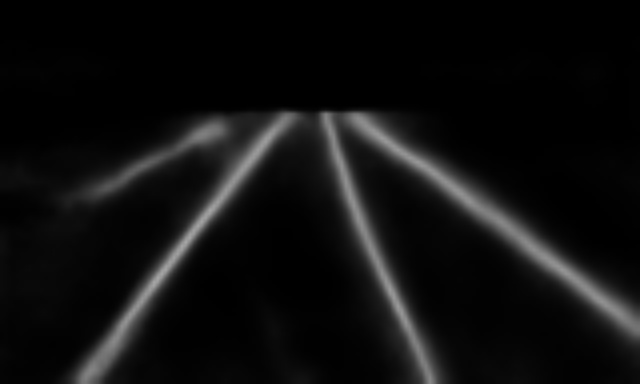}}\,\!\!
    \subfloat {\includegraphics[width=1.7cm,height=1.1cm]{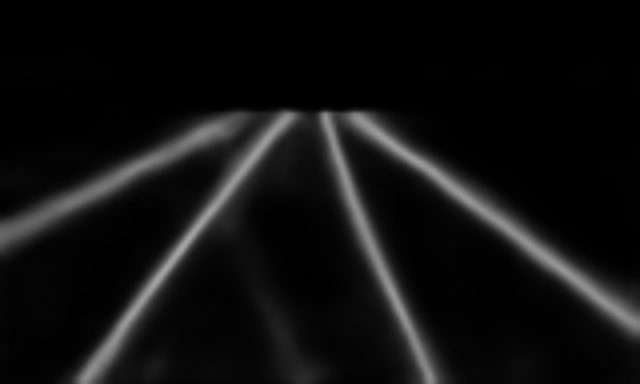}}\,\!\!
    \subfloat {\includegraphics[width=1.7cm,height=1.1cm]{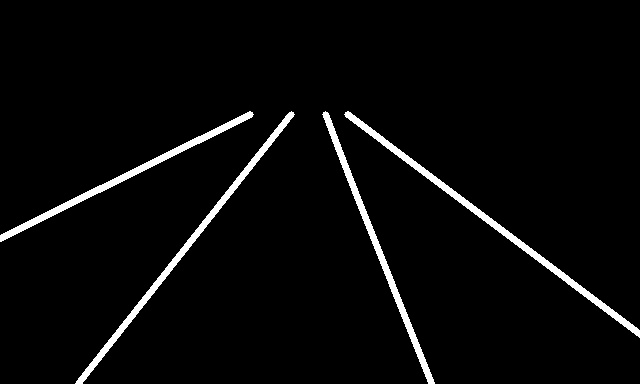}}\\

    \setcounter{subfigure}{-1}
    \subfloat[$I_t$] {\includegraphics[width=1.7cm,height=1.1cm]{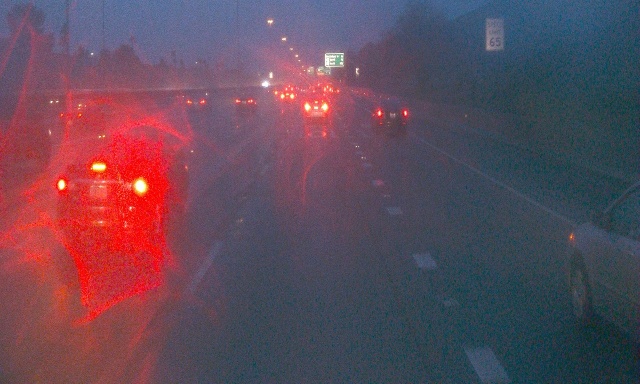}}\,\!\!
    \subfloat[$\tilde{O}_t$] {\includegraphics[width=1.7cm,height=1.1cm]{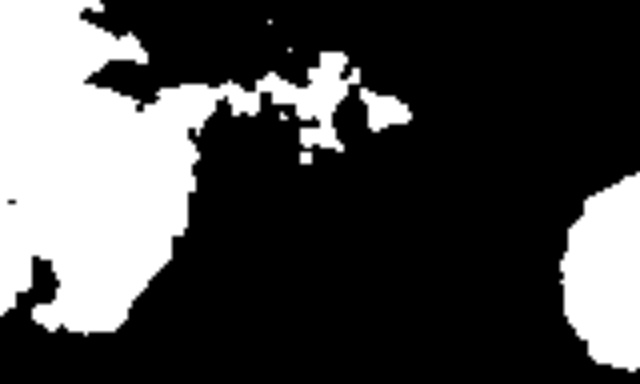}}\,\!\!
    \subfloat[$\tilde{F}_t$] {\includegraphics[width=1.7cm,height=1.1cm]{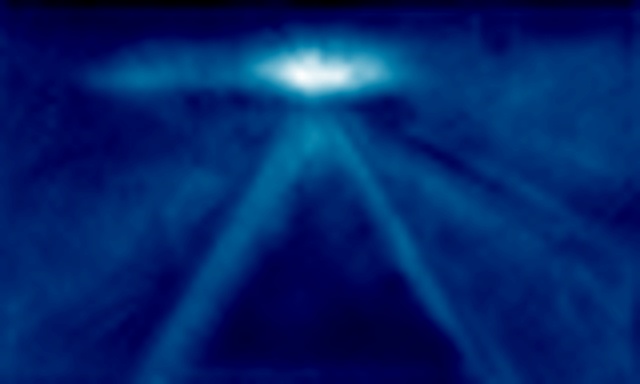}}\,\!\!
    \subfloat[$F_t$] {\includegraphics[width=1.7cm,height=1.1cm]{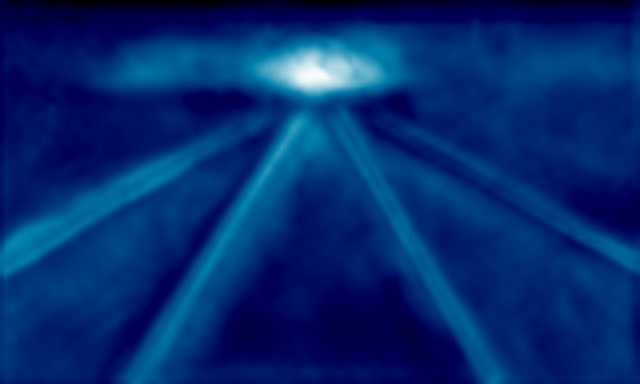}}\,\!\!
    \subfloat[$\tilde{P}_t$] {\includegraphics[width=1.7cm,height=1.1cm]{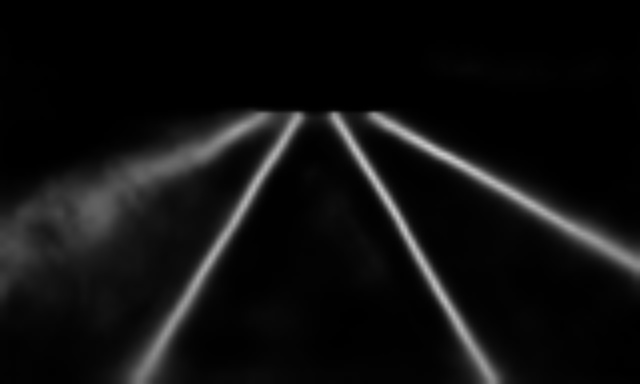}}\,\!\!
    \subfloat[$P_t$] {\includegraphics[width=1.7cm,height=1.1cm]{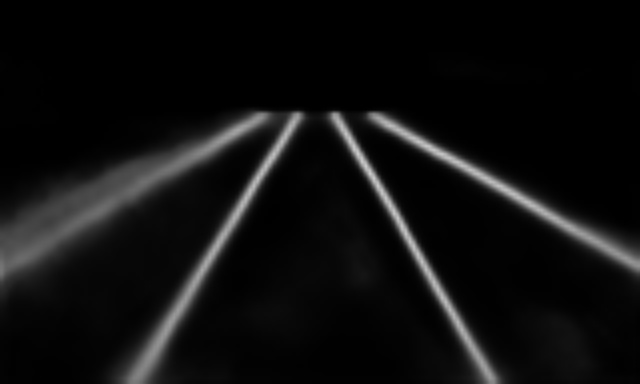}}\,\!\!
    \subfloat[Ground-truth] {\includegraphics[width=1.7cm,height=1.1cm]{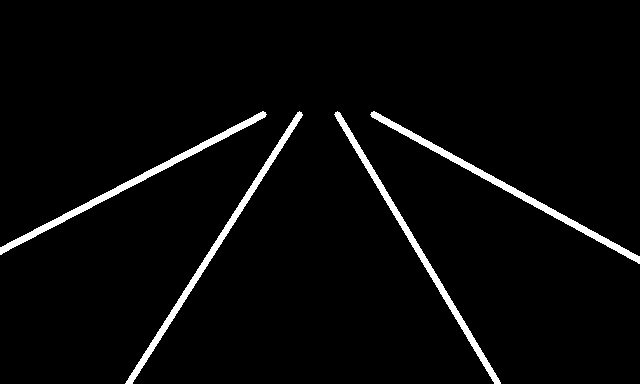}}

    \setcounter{subfigure}{-1}
    \caption
    {
         Visualization of the obstacle mask $\tilde{O}_t$, the feature map $\tilde{F}_t$, and the probability map $\tilde{P}_t$, and their enhanced ones.
    }
    \label{fig:activation2}
    \end{center}

\end{figure}

\noindent\textbf{Efficacy of obstacle mask:} As compared with the proposed algorithm (Method \RomNum{4}), Method~\RomNum{1} yields inferior scores in terms of $\rm F1$ and ${\rm R}_{\rm M}$. This indicates that utilizing obstacle masks is beneficial for accurate and temporally stable lane detection. Some detected obstacle masks are presented in the second column in Fig.~\ref{fig:activation2}.

\noindent\textbf{Efficacy of memory information:}
Compared to Method \RomNum{4}, Method \RomNum{2} yields slightly lower scores of ${\rm R}_{\rm F}$, but its F1 score and missing rate ${\rm R}_{\rm M}$ become worse. This indicates that, rather than using previous output only, it is more effective to exploit memory information for reliable lane detection.

\noindent\textbf{Efficacy of synthetic training data:}
Without using synthetic data in Method \RomNum{3}, the performances drop significantly in every metric, especially for ${\rm R}_{\rm F}$ and ${\rm R}_{\rm M}$. The synthetic data augmentation is helpful for enhancing the temporal stability of lanes.

\noindent\textbf{Efficacy of OMR module:}
Fig.~\ref{fig:activation2} visualizes the obstacle mask $\tilde{O}_t$, the feature map $\tilde{F}_t$ and the lane probability map $\tilde{P}_t$ of a current frame $I_t$, and their refined ones $F_t$ and $P_t$. Some lane parts in $I_t$ are occluded by nearby obstacles, and thus their features and probabilities are erroneous. These results, however, are restored faithfully through the proposed OMR module.

\captionsetup[subfigure]{labelformat=empty}

\noindent\textbf{Runtime:} Table~\ref{table:runtime} lists the runtime for each stage of the proposed algorithm. The processing speed is about 105 frames per second, surpassing RVLD \cite{jin2023}. RVLD demands high computational costs due to local correlation in the motion estimator. In contrast, the proposed algorithm consists of simpler operations. The key parts, the latent obstacle detection and OMR, take less time to process. The decoding part requires the longest time, containing the NMS process.

\begin{table*}[t]\centering
    \renewcommand{\arraystretch}{0.85}
    \caption
    {
        Runtime analysis and comparison of the proposed algorithm with RVLD. LOD refers to latent obstacle detection. The processing times are reported in seconds per frame.
    }

    \resizebox{0.85\linewidth}{!}{
    \begin{tabular}[t]{C{2.0cm}^C{2.0cm}^C{2.0cm}^C{2.0cm}^C{2.0cm}^C{2.0cm}}
    \toprule
    \multirow{2}{*}{RVLD \cite{jin2023}} & \multicolumn{5}{c}{Proposed}\\
    \cmidrule(lr){2-6}
     & Encoding & LOD & OMR & Decoding & Total\\
    \midrule
        0.0125s & 0.0026s & 0.0002s & 0.0010s & 0.0057s & 0.0095s \\

    \bottomrule
    \end{tabular}}
    \label{table:runtime}

\end{table*}

\section{Conclusions}
We proposed a novel video lane detector. First, the proposed algorithm extracts a feature map for a current frame and detects latent obstacles obstructing lane visibility. Then, it enhances the feature map using the occlusion-aware memory-based refinement (OMR) module, which takes the detected obstacle mask and the feature map from the current frame, the previous output, and the memory information as input. The enhanced feature map is used for more reliable lane detection. Moreover, we developed a data augmentation scheme for training the OMR module robustly. Experimental results demonstrated that the proposed algorithm outperforms existing techniques meaningfully.

\section*{Acknowledgements}

This work was conducted by Center for Applied Research in Artificial Intelligence (CARAI) grant funded by DAPA and ADD (UD230017TD) and was supported by the National Research Foundation of Korea (NRF) grants funded by the Korea government (MSIT) (No.~NRF-2022R1A2B5B03002310 and No.~RS-2024-00397293).

\clearpage

%
%
\bibliographystyle{splncs04}
\bibliography{04822}
\end{document}